%% file: main.tex
\documentclass{article}

    \usepackage[nonatbib,final]{neurips_2020}

\usepackage[utf8]{inputenc} % allow utf-8 input
\usepackage[T1]{fontenc}    % use 8-bit T1 fonts
\usepackage{hyperref}       % hyperlinks
\usepackage{url}            % simple URL typesetting
\usepackage{booktabs}       % professional-quality tables
\usepackage{amsfonts}       % blackboard math symbols
\usepackage{nicefrac}       % compact symbols for 1/2, etc.
\usepackage{microtype}      % microtypography
\usepackage{glossaries}
\usepackage{graphicx}
\usepackage{todonotes}
\usepackage[frozencache]{minted}
\usepackage{listings}
\usepackage{subcaption}
\usepackage{wrapfig}
\usepackage{caption}

\setminted[python]{
	linenos,
	tabsize=4,
	frame=lines,
	breaklines,
	fontsize=\small
}

\loadglsentries{glossary}
\glsdisablehyper

\title{PyTorch-Hebbian: facilitating local learning in a deep learning framework}

\author{%
  Jules Talloen\thanks{\url{www.ml6.eu}, \url{www.julestalloen.eu}} \\
  Ghent University / ML6\\
  \texttt{jules@talloen.eu} \\
  \And
  Joni Dambre \hspace{40pt} Alexander Vandesompele \\
  Ghent University - imec \\
  \texttt{\{joni.dambre, alexander.vandesompele\}@ugent.be}
}

\begin{document}
\nolinenumbers

\maketitle

\begin{abstract}
  Recently, unsupervised local learning, based on Hebb's idea that change in synaptic efficacy depends on the activity of the pre- and postsynaptic neuron only, has shown potential as an alternative training mechanism to backpropagation. Unfortunately, Hebbian learning remains experimental and rarely makes it way into standard deep learning frameworks. In this work, we investigate the potential of Hebbian learning in the context of standard deep learning workflows. To this end, a framework for thorough and systematic evaluation of local learning rules in existing deep learning pipelines is proposed. Using this framework, the potential of Hebbian learned feature extractors for image classification is illustrated. In particular, the framework is used to expand the Krotov-Hopfield learning rule to standard convolutional neural networks without sacrificing accuracy compared to end-to-end backpropagation. The source code is available at \url{https://github.com/Joxis/pytorch-hebbian}.
\end{abstract}

\section{Introduction}
In this work, we study unsupervised local learning rules which rely solely on bottom-up information propagation. Furthermore, we limit the analysis to Hebbian learning by imposing correlated pre- and postsynaptic activity as a synaptic reinforcement requirement \cite{hebb_organization_1949}. A variety of Hebbian plasticity-based learning rules for neural networks have been proposed \cite{pehlevanHebbianAntiHebbianNeural2015,jacynaClassificationPerformanceHopfield1989,bahroun_building_2017, krotov_unsupervised_2019, grinberg_local_2019}. In the remainder of this work, we will mainly focus on the Krotov-Hopfield learning rule \cite{krotov_unsupervised_2019} due to its remarkable performance despite the locality constraint. The learning rule will be used to train the first layer(s) of a neural network for image feature extraction. Next, using these features, backpropagation is used to train the output layer to detect image classes. 

Deep learning with backpropagation was greatly accelerated by frameworks such as TensorFlow \cite{noauthor_tensorflow_nodate} and PyTorch \cite{noauthor_pytorch_nodate}. In particular, high level abstractions built on top of these frameworks, like Keras \cite{cholletKerasPythonDeep}, allow rapid experimentation without worrying about the underlying complexity. Unfortunately, the Hebbian research community lacks such established tools.
Because of their focus on gradient based optimization, standard deep learning frameworks are not directly suitable for Hebbian learning. Nonetheless, these frameworks offer many useful features, independent of the underlying training mechanism. For this reason, an entirely new framework for Hebbian learning would be redundant and backward incompatible. A new paradigm, making use of established functionality, is preferable. 

This work proposes a novel PyTorch framework for Hebbian learning. First, the objectives for the framework are set out, in Section \ref{sec:objectives}. Next, in Section \ref{sec:framework}, the framework and its deviations from the classic paradigm are elucidated. To conclude, the performance of the framework with the Krotov-Hopfield \cite{krotov_unsupervised_2019} learning rule is illustrated, in Section \ref{sec:results}.

\section{Related work} \label{sec:related work}
Few implementations are published and those that are, are typically limited. These implementations are either written based on deprecated libraries \cite{ironbarIronbarTheanoGeneralized2018}, limited to a specific algorithm \cite{kornilovMatweyPythongha2020} or do not adhere to a conventional deep learning workflow scheme \cite{gabrielelaganiGabrieleLaganiHebbianLearningThesis2020}.
At the time of writing, the most starred implementation of a Hebbian learning algorithm on GitHub is by Raphael Holca \cite{raphaelholcaRaphaelholcaHebbianRL2020, raphaelholcaRaphaelholcaHebbianCNN2020}. He wrote Python code for a type of Hebbian and reward-based learning in deep \glspl{cnn}.

\section{Objectives} \label{sec:objectives}
Existing Hebbian learning implementations typically do not support easy integration with standard deep learning workflows. As a consequence, these implementations are rarely reused and valuable time is wasted. The goal of this work is to introduce a novel framework for thorough and systematic evaluation of Hebbian learning in the context of a standard deep learning execution flow. Although created with Hebbian learning in mind, it is generally applicable to unsupervised local feature learning. The framework is based on PyTorch Ignite \cite{noauthor_pytorch-ignite_nodate}, a high level, yet flexible, library to aid in training neural networks.

We use established tools as much as possible. Instead of building yet another framework, various modular components are structured and combined to take advantage of existing work. Even though Hebbian learning differs from standard deep learning in many aspects, it is implemented making use of most of PyTorch's standard components. This allows the use of existing PyTorch models with little to no changes.

\section{Framework} \label{sec:framework}
\begin{wrapfigure}[26]{r}{0.5\textwidth}
	\centering
	\includegraphics[width=.48\textwidth]{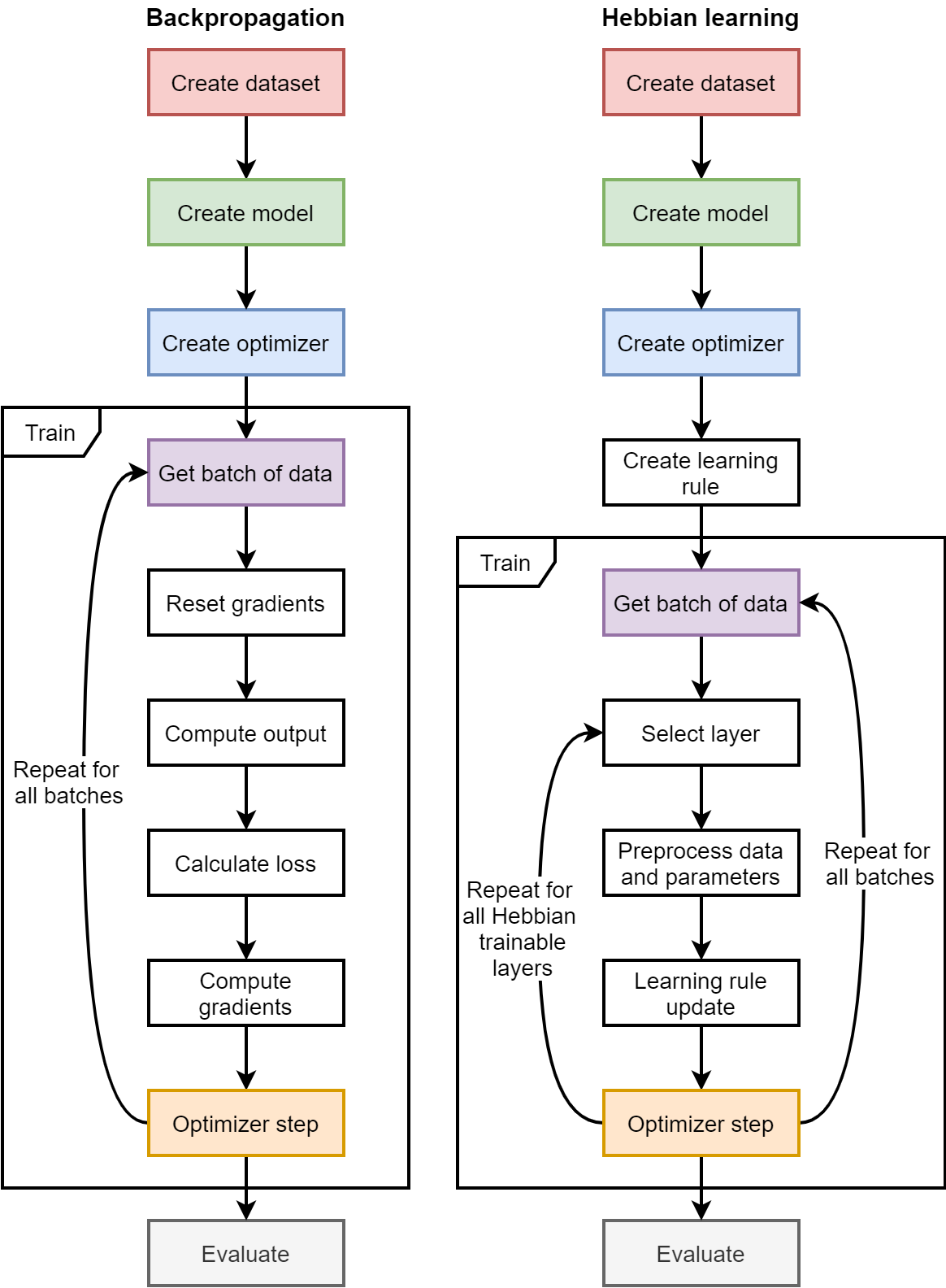}
	\caption{A comparison of the execution flow of backpropagation and Hebbian learning. A single epoch is shown.}
	\label{fig:hebb_vs_sup_flow}
\end{wrapfigure}

Hebbian learning introduces a new learning paradigm with an accompanying shift in execution flow. The flow for a single epoch, compared to the classical backpropagation execution flow (in PyTorch), is illustrated in Figure \ref{fig:hebb_vs_sup_flow}. For Hebbian learning, an additional learning rule is required. Furthermore, there is no single forward pass and the Hebbian trainable layers are trained separately. The Hebbian paradigm thus introduces a second loop, iterating over the layers.

The core architecture of the framework is based on PyTorch Ignite's \cite{noauthor_pytorch-ignite_nodate} \texttt{Engine} and event system. Two new classes were introduced: the \texttt{Trainer} and \texttt{Evaluator}. By attaching event \texttt{Handlers} to the \texttt{Trainer} for one or more \texttt{Events}, various powerful functionalities can be achieved. A \texttt{Handler} can be anything from an \texttt{Evaluator} to a \texttt{Visualizer}.

A minimal code example of the framework and its components is given in Listing \ref{lst:framework}. First, the model, dataset and dataloader are created. Next, the novel learning rule, optimizer and trainer are created. Hebbian learning is now possible within PyTorch with only a few lines of code.

\begin{listing}[!ht]
\begin{minted}{python}
# Create the model, dataset and dataloader
model = create_model()
dataset = MNIST(root=config.DATASETS_DIR, transform=transforms.ToTensor())
train_loader = DataLoader(dataset, batch_size=64, shuffle=True)

# Create the learning rule, optimizer and trainer
learning_rule = KrotovsRule()
optimizer = Local(model.named_parameters(), lr=0.01)
trainer = HebbianTrainer(model, learning_rule, optimizer)

# Run the trainer
trainer.run(train_loader, epochs=10)
\end{minted}
	\caption{A minimal example of the Hebbian framework. First, create a model, dataset and dataloader. Next, create the learning rule, optimizer and trainer. Finally, run the trainer.}
	\label{lst:framework}
\end{listing}

\subsection{Hebbian trainer}
The \texttt{Trainer} is the essence of the framework and is divided into two subclasses: \texttt{SupervisedTrainer} and \texttt{HebbianTrainer}. The \texttt{SupervisedTrainer} is a simple abstraction on top of a typical supervised training loop whereas the \texttt{HebbianTrainer} introduces a novel paradigm for local learning. The \texttt{HebbianTrainer} does not update all weights at once for each batch. Instead, a new loop iterating over the layers to perform local weight updates, is introduced. The current implementation updates these layers in order but parallel updates, based on the weights of the previous iteration, are also possible. As a result, Hebbian learning only requires a single forward pass per trainable layer.

The \texttt{HebbianTrainer} accepts multiple learning rules, allowing a different learning rule to be used per layer. In addition, the trainer takes an optional \texttt{supervised\_from} and \texttt{freeze\_layers} argument. The first specifies from which layer on to start supervised training and the latter expects a list of layer names to exclude from training. These parameters allow for flexibility in the training procedure.

Because of its layer-wise weight updates, the \texttt{HebbianTrainer} requires independent access to each layer's parameters. To this end, the novel \texttt{Local} optimizer accepts a dictionary, mapping layer names to parameters, as input. Now, when the learning rule has computed the weight update for a specific layer, it is possible to signal the optimizer to only update the corresponding parameters.

\subsection{Hebbian evaluator} \label{subsec:hebb evaluator}
The \texttt{HebbianEvaluator} comprises of both a \texttt{SupervisedTrainer} and \texttt{SupervisedEvaluator}. A model is only trained with Hebbian learning rules up until a certain layer. From this layer on, the \texttt{SupervisedTrainer} takes over in order to achieve task dependent outputs. The supervised layers are optimized based on the current state of the Hebbian trained layers. Once these layers are trained, the entire model is evaluated with a \texttt{SupervisedEvaluator} (See Listing \ref{lst:evaluator}). 

\begin{listing}[!htb]
\begin{minted}{python}
evaluator = SupervisedEvaluator(model, criterion)
evaluator.attach(trainer, Events.EPOCH_COMPLETED, val_loader)
\end{minted}
	\caption{An example showing how to attach an \texttt{Evaluator} to a \texttt{Trainer}. The evaluator will run every epoch on the \texttt{EPOCH\_COMPLETED} event of the trainer.}
	\label{lst:evaluator}
\end{listing}

\subsection{Visualizers and loggers}
Hebbian learning by itself lacks clear performance metrics to track the training progress. Only by fully training the final layer(s) with \gls{backpropagation}, one can assess the performance of the Hebbian trained layers. For this reason it is useful to closely track the model to have a better understanding of what is going on during training.
The proposed framework supports metric and training progress logging through various channels. Both textual interfaces, such as the command line, and more visual interfaces, such as TensorBoard \cite{noauthor_tensorboard_nodate}, are supported. 

\begin{figure}[!htb]
	\centering
	\begin{subfigure}{.42\textwidth}
		\centering
		\includegraphics[width=.75\textwidth]{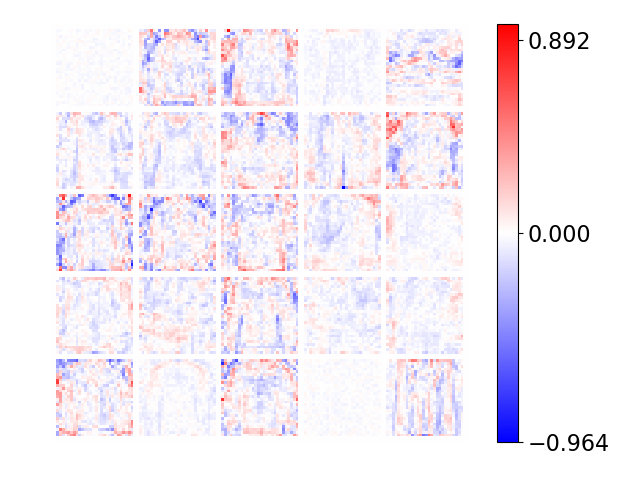}
		\caption{}
		\label{subfig:mnist-fashion-weights_comparison-backprop}
	\end{subfigure}
	\begin{subfigure}{.42\textwidth}
		\centering
		\includegraphics[width=.75\textwidth]{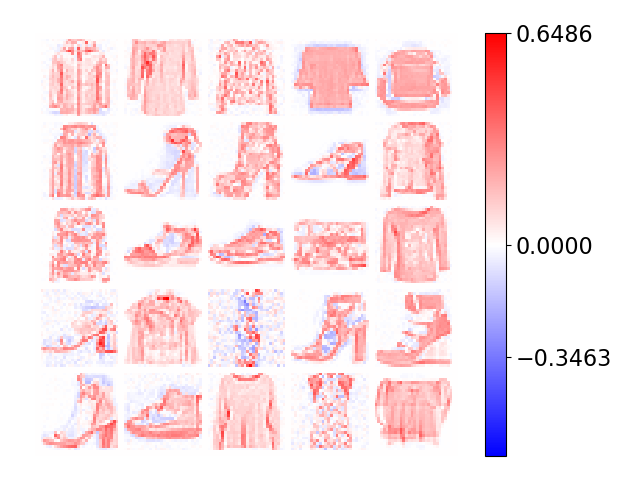}
		\caption{}
		\label{subfig:mnist-fashion-weights_comparison-hebbian}
	\end{subfigure}
	\caption{The incoming weights of 25 (out of 2000) randomly sampled fully connected hidden units trained end-to-end with backpropagation (\subref{subfig:mnist-fashion-weights_comparison-backprop}) and the Krotov-Hopfield learning rule (\subref{subfig:mnist-fashion-weights_comparison-hebbian}) on the MNIST fashion dataset.}
	\label{fig:mnist-fashion-weights_comparison}
\end{figure}

Various \texttt{Visualizers}, including a weights visualizer, unit convergence graph, weights histogram and activation statistics were added. An example of the \texttt{WeightsImageHandler} for 25 linear units, trained with backpropagation (\subref{subfig:mnist-fashion-weights_comparison-backprop}) and the Krotov-Hopfield learning rule (\subref{subfig:mnist-fashion-weights_comparison-hebbian}), is shown in Figure \ref{fig:mnist-fashion-weights_comparison}.

\section{Results} \label{sec:results}
Using the novel framework, the convolutional layer of a \gls{cnn} with 400 filters was trained with the Krotov-Hopfield learning rule \cite{krotov_unsupervised_2019} on the MNIST fashion dataset \cite{xiao_fashion-mnist_2017}.\footnote{See the supplementary material for a detailed description of the experiment setup.} The output layer was then trained with backpropagation. As shown in Table \ref{tab:reproduction results-fashion-conv}, the resulting network achieves 91.44\% test accuracy, which is only 0.5\% lower compared to the same network trained end-to-end with backpropagation.

\begin{table}[!htb]
    \caption{Summary of the train and test accuracy of the \gls{cnn} on the MNIST fashion dataset. Hebbian learning followed by backpropagation compared to end-to-end backpropagation.}
	\label{tab:reproduction results-fashion-conv}
	\centering
	\begin{tabular}{llll}
	    \toprule
		& Hebb + backprop & End-to-end backprop \\
		\midrule
		train accuracy (\%) & 98.64 & 99.98 \\
		\midrule
		test accuracy (\%) & 91.44 & 91.94 \\
		\bottomrule
	\end{tabular}
\end{table}

Building upon PyTorch's tensor backend, the proposed framework fully supports the \gls{cuda} \cite{noauthor_cuda_2017}. Moreover, the local character of the Hebbian learning paradigm offers more room for parallelism compared to \gls{backpropagation}. For the Krotov-Hopfield learning rule \cite{krotov_unsupervised_2019} on the MNIST digits dataset \cite{lecun_mnist_nodate}, our proposed framework is 14 times faster with only 3 seconds per epoch compared to 44 seconds for the original numpy \cite{harris2020array} implementation \cite{krotov_dimakrotovbiological_learning_2020}.\footnote{See the supplementary material for a detailed description of the used hardware and experiment setup.}

% \begin{table}[!htb]
%     \caption{Time (in seconds) required per epoch on the MNIST digits dataset \cite{lecun_mnist_nodate} for the numpy implementation (CPU only) by Krotov et al. \cite{krotov_dimakrotovbiological_learning_2020} compared to this work (GPU accelerated). The proposed framework is more than 14 times faster.}
% 	\label{tab:framework performance}
% 	\centering
% 	\begin{tabular}{lll}
% 	    \toprule
% 		& Krotov et al. (CPU only) & This work (GPU accelerated) \\
% 		\midrule
% 		Time (seconds/epoch) & 44 & 3 \\
% 		\bottomrule
% 	\end{tabular}
% \end{table}

While the proposed framework offers significant improvements, there is still room for further parallelization. Grinberg et al. built a fast C++ library for \gls{cuda} that takes most advantage of the concept of local learning and optimizes performance \cite{grinberg_local_2019}. However, their implementation is not compatible with existing machine learning frameworks and imposes specific feature and bandwidth requirements on the hardware. In comparison, the framework proposed in this work is PyTorch-compatible and does not require any special hardware. The implementation by Grinberg et al. is not available and an exact performance comparison is thus not possible.

\section{Conclusion} \label{sec:conclusion}
In this work we have shown that Hebbian learning can be integrated into standard deep learning workflows. Furthermore, we proposed a novel PyTorch framework to easily do so. Using this framework, existing PyTorch models can be Hebbian trained and evaluated with minimal effort. Additionally, this work shows that Hebbian learned features are able to capture useful information to achieve similar classification accuracy, on MNIST fashion, compared to end-to-end backpropagation. Whether similar results can be obtained on more challenging datasets, such as ImageNet \cite{imagenet_cvpr09}, remains unknown. Lastly, Hebbian learning exposes additional potential with regards to computational improvements via parallelization. 

\begin{ack}
% \todo[inline]{Use unnumbered first level headings for the acknowledgments. All acknowledgments
% More information about this disclosure can be found at: \url{https://neurips.cc/Conferences/2020/PaperInformation/FundingDisclosure}.}
This work originated from a master's dissertation of the computer science engineering program at Ghent University, under the supervision of Joni Dambre and Alexander Vandesompele. The work was later completed with the support of ML6. Special thanks to Jan Van Looy, Thomas Janssens and Matthias Feys.
\end{ack}

\bibliographystyle{ieeetr}
\bibliography{main}{}

\end{document}

% \nolinenumbers

\maketitle

\section{Computational performance comparison}
The proposed framework is more than 14 times faster compared to the original numpy \cite{harris2020array} implementation \cite{krotov_dimakrotovbiological_learning_2020} of the Krotov-Hopfield learning rule \cite{krotov_unsupervised_2019}. The latter implementation only supports \glspl{cpu}. The comparison is made in Table \ref{tab:framework performance}. The experiments were conducted on an Intel Core i7 4702HQ \gls{cpu} and a NVIDIA Tesla P100 \gls{gpu}. Both devices had sufficient memory. Hebbian learning, with the Krotov-Hopfield learning rule, was performed on the train set of MNIST digits (60,000 samples) with a batch size of 1024. The network is fully connected with one hidden layer, consisting of 2000 hidden units. An average number of seconds per epoch was taken over 10 epochs.

\begin{table}[!htb]
  \caption{Time (in seconds) required per epoch for the numpy implementation (CPU only) by Krotov et al. \cite{krotov_dimakrotovbiological_learning_2020} compared to this work (GPU accelerated). The proposed framework is more than 14 times faster.}
  \label{tab:framework performance}
  \centering
  \begin{tabular}{lll}
    \toprule
     & Krotov et al. (CPU only) & This work (GPU accelerated) \\
    \midrule
    Time (seconds/epoch) & 44 & 3 \\
    \bottomrule
  \end{tabular}
\end{table}

\section{Single-layer CNN image classification}
In this experiment the proposed framework is used to evaluate a single-layer \gls{cnn} for image classification on the MNIST fashion dataset \cite{xiao_fashion-mnist_2017}.

\begin{figure}[!htb]
	\centering
	\includegraphics[width=0.25\textwidth]{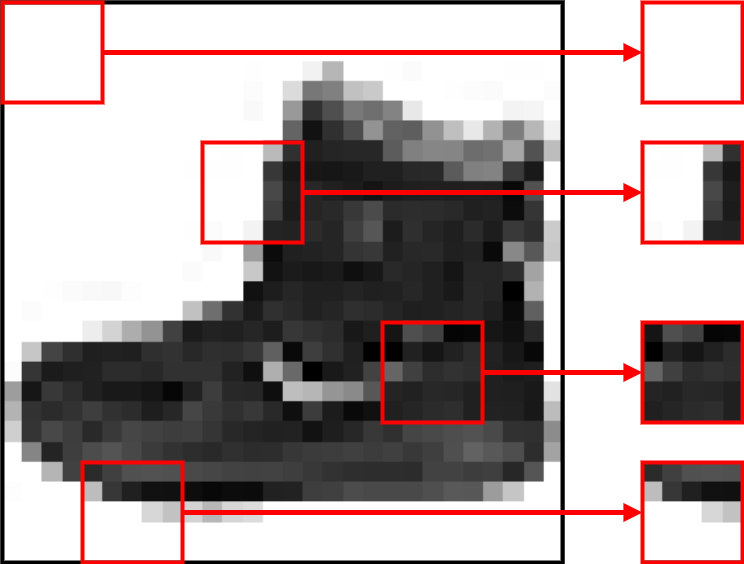}
	\caption{Image patches extracted from an MNIST fashion image. Four $5 \times 5$ patches, extracted from four locations, are shown.}
	\label{fig:image_patch_preprocess}
\end{figure}

The Krotov-Hopfield learning rule \cite{krotov_unsupervised_2019} can not simply be fed the inputs and layer weights from a \gls{cnn} in their current form. The dimensions of the input image and the convolution filters do not match and the learning rule expects these to be the same. A preprocessing function precedes the learning rule update function. This function prepares the inputs so that the learning rule can work layer-agnostic. For a convolutional layer, the preprocessing step extracts patches from the input image. This is done similarly to how the convolution kernel slides over the image. For each kernel position, the input pixels at that position are extracted to form a new image, or image patch. The process is sketched in Figure \ref{fig:image_patch_preprocess}, for a $5 \times 5$ kernel. The red squares represent the kernel and the extracted image patch is shown for four kernel positions. 
After the preprocessing step, the learning rule inputs are of the same dimensions as the kernel and Hebbian updates can be calculated. Each convolution filter is treated as a hidden unit and all these filters are ranked per batch of image patches. 

Simply put, the convolution filters are trained on all image patches of a given dimension from all input images. \Glspl{cnn} are very powerful because of the scale of this set of image patches. The filters are trained on a lot more training samples compared to a single hidden unit in a fully connected network.

\subsection{Model} \label{subsubsec:conv models} 
\begin{figure}[!htb]
	\centering
	\includegraphics[width=0.55\textwidth]{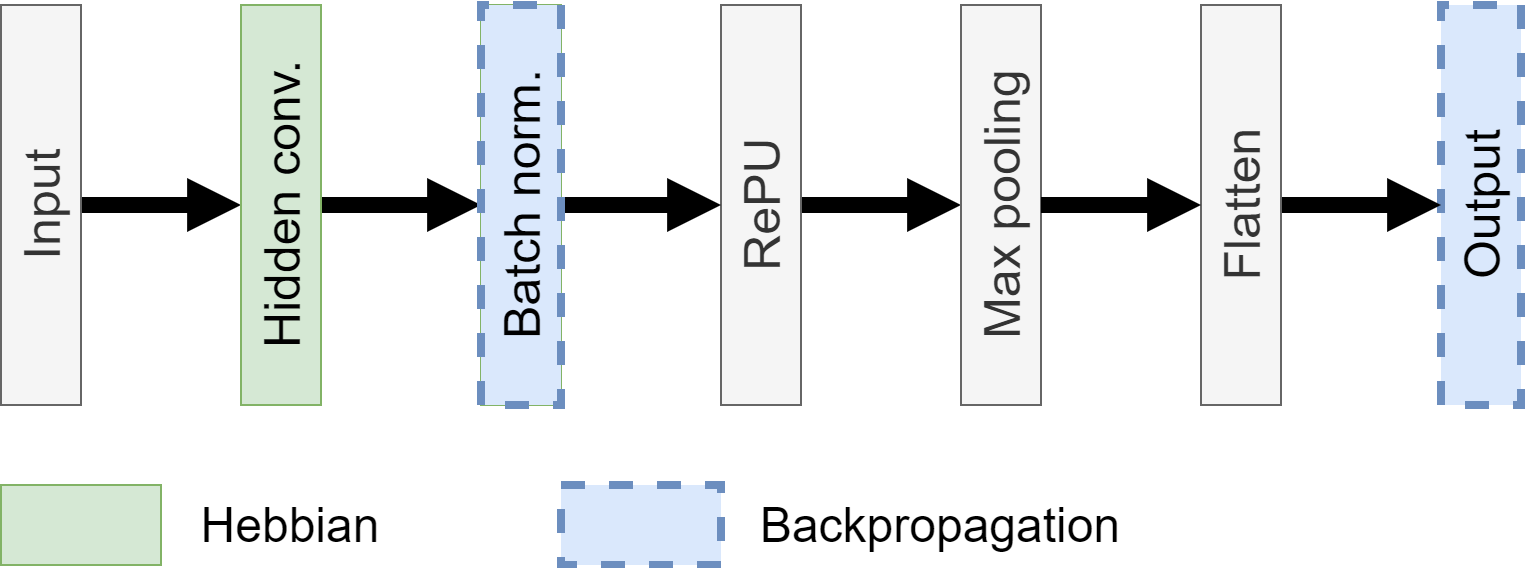}
	\caption{The layers of the \gls{cnn} with a single hidden convolutional layer. The convolutional layer is Hebbian trained. The batch normalization and output layer are trained with backpropagation. The other layers do not contain trainable parameters.}
	\label{fig:layers_batchnorm-conv}
\end{figure}

The layers are shown in Figure \ref{fig:layers_batchnorm-conv}. The model receives a $28\times 28$ image with $1$ color channel as input. Next is a convolutional layer with $N=400$ filters, followed by a batch normalization, \gls{repu} and max pooling layer. Finally, the network flattens the convolutional features to a single dimension, before passing them to the fully connected output layer. The output layer has $K=10$ units, one for each class. The convolutional layer has $5 \times 5$ kernels with stride 1 and the pooling layer has $2 \times 2$ kernels with stride 2.

\subsection{Setup} \label{subsec:conv general setup}
The training setup is described by the following steps:
\begin{enumerate}
	\item The Hebbian trainable layer(s) are standard normal initialized.
	\item The Hebbian layer(s) are trained with the Krotov-Hopfield learning rule \cite{krotov_unsupervised_2019} and \texttt{Local} optimizer.
	\subitem During training, the learning rate is linearly decreased to 0  at the final epoch.
	\item The Hebbian layer(s) are frozen.
	\item The final layer is trained with backpropagation using cross-entropy loss and the Adam optimizer.
	\subitem The learning rate is scheduled by a \texttt{ReduceLROnPlateau} handler.
	\subitem Early stopping is used to prevent overfitting.
\end{enumerate}

\paragraph{Hyperparameters}
The hyperparameters were manually selected and tuned based on the insights from previous work. Hebbian learning was performed for only 100 epochs to limit the time required per experiment. The resulting hyperparameters are summarized below:
\begin{itemize}
	\item \textbf{unsupervised batch size}: 1000 patches
	\item \textbf{unsupervised epochs}: 100
	\item \textbf{unsupervised learning rate}: 0.04
	\item \textbf{$\boldsymbol{\Delta}$}: 0.4
	\item \textbf{$\boldsymbol{k}$}: 4
	\item \textbf{$\boldsymbol{p}$}: 4
	\item $\boldsymbol{n}$: 1
	\item \textbf{supervised batch size}: 256
	\item \textbf{supervised epochs}: 100
	\item \textbf{supervised learning rate}: 0.001
\end{itemize}

\subsection{Results}
\begin{figure}[!htb]
	\centering
	\includegraphics[width=0.55\textwidth]{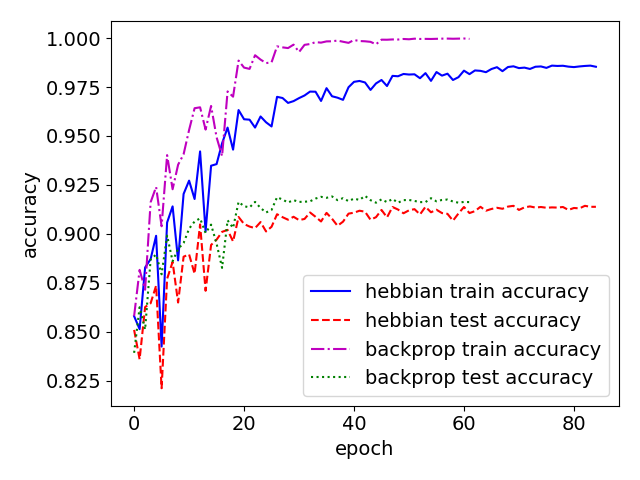}
	\caption{Train and test accuracy curves for the \gls{cnn} trained on MNIST fashion with end-to-end backpropagation compared to backpropagation for the final layer only, with a Hebbian learned hidden layer.}
	\label{fig:mnist-fashion-conv-histories}
\end{figure}

The train and test curves are shown in Figure \ref{fig:mnist-fashion-conv-histories}. With only 400 convolution filters the network achieves 91.44\% test accuracy, which is only 0.5\% lower compared to the same network trained end-to-end with backpropagation. The results are summarized in Table \ref{tab:reproduction results-fashion-conv}. The model trained end-to-end with backpropagation converges faster. However, it is also prone to overfitting, as can be seen by the 'backprop test accuracy' curve in Figure \ref{fig:mnist-fashion-conv-histories}. While the training accuracy continues to approach 100\%, the test accuracy starts slightly dropping.

\begin{table}[!htb]
    \caption{Summary of the train and test accuracy of the \gls{cnn} on the MNIST fashion dataset. Hebbian learning followed by backpropagation compared to end-to-end backpropagation.}
	\label{tab:reproduction results-fashion-conv}
	\centering
	\begin{tabular}{llll}
	    \toprule
		& Hebb + backprop & End-to-end backprop \\
		\midrule
		train accuracy (\%) & 98.64 & 99.98 \\
		\midrule
		test accuracy (\%) & 91.44 & 91.94 \\
		\bottomrule
	\end{tabular}
\end{table}

\bibliographystyle{ieeetr}
\bibliography{main}{}

%% file: glossary.tex
% Acronyms
\makeglossaries
\newacronym{ml}{ML}{machine learning}
\newacronym{ai}{AI}{artificial intelligence}
\newacronym{ann}{ANN}{artificial neural network}
\newacronym{snn}{SNN}{spiking neural network}
\newacronym{cnn}{CNN}{convolutional neural network}
\newacronym{ltp}{LTP}{long-term potentiation}
\newacronym{ltd}{LTD}{long-term depression}
\newacronym{pca}{PCA}{principal component analysis\glsadd{gls-pca}}
\newacronym{psp}{PSP}{postsynaptic potential}
\newacronym{epsp}{EPSP}{excitatory postsynaptic potential}
\newacronym{ipsp}{IPSP}{inhibitory postsynaptic potential}
\newacronym{stdp}{STDP}{spike-timing-dependent plasticity}
\newacronym{rnn}{RNN}{recurrent neural networks}
\newacronym{ad}{AD}{automatic differentiation}
\newacronym{gpu}{GPU}{graphics processing unit}
\newacronym{cpu}{CPU}{central processing unit}
\newacronym{cuda}{CUDA\textsuperscript{\tiny\textregistered}}{compute unified device architecture}
\newacronym{cli}{CLI}{command line interface}
\newacronym{uml}{UML}{unified modeling language}
\newacronym{gmdh}{GMDH}{group method of data handling}
\newacronym{relu}{ReLU}{rectified linear unit}
\newacronym{repu}{RePU}{rectified polynomial unit}
\newacronym{dam}{DAM}{dense associative memory}
\newacronym{lif}{LIF}{leaky integrate-and-fire}
\newacronym{mpc}{MPC}{mixture of principal components}
\newacronym{elm}{ELM}{extreme learning machine}
\newacronym{som}{SOM}{self-organizing map}

% Terms
\newglossaryentry{learning rule}
{
	name=learning rule,
	description={A mechanism modelling learning in a neural network}
}
\newglossaryentry{receptive field}
{
	name=receptive field,
	description={The portion of sensory input that can elicit neuronal responses when stimulated \cite{alonsoReceptiveField2009}}
}
\newglossaryentry{feature}
{
	name=feature,
	description={In machine learning and pattern recognition, a feature is an individual measurable property or characteristic of a phenomenon being observed \cite{bishopPatternRecognitionMachine2006}}
}
\newglossaryentry{gls-pca}
{
	name=principal component analysis,
	description={Multivariate technique that analyzes a data table in which observations are described by several inter‐correlated quantitative dependent variables \cite{abdiPrincipalComponentAnalysis2010}. Its goal is to extract the important information from the table, to represent it as a set of new orthogonal variables called principal components}
}
\newglossaryentry{lateral inhibition}
{
	name=lateral inhibition,
	description={The capacity of an activated neuron to inhibit the activity of other neurons}
}
\newglossaryentry{backpropagation}
{
	name=backpropagation,
	description={Backpropagation repeatedly adjusts the weights of the connections in a network so as to minimize a loss function of the difference between the actual output vector of the network and the desired output vector. All weights are updated via gradient descent: the gradient of the loss function with respect to the network's weights is calculated and backwards propagated by iterating the chain rule. See Section \ref{sec:backpropagation}}
}